\documentclass[lettersize,journal]{IEEEtran}
\usepackage{amsmath,amsfonts}
\usepackage{algorithmic}
\usepackage{algorithm}
\usepackage{array}
\usepackage[caption=false,font=normalsize,labelfont=sf,textfont=sf]{subfig}
\usepackage{textcomp}
\usepackage{stfloats}
\usepackage{url}
\usepackage{verbatim}
\usepackage{graphicx}
\def\BibTeX{{\rm B\kern-.05em{\sc i\kern-.025em b}\kern-.08em
    T\kern-.1667em\lower.7ex\hbox{E}\kern-.125emX}}
\usepackage{balance}

\usepackage{booktabs}
\usepackage{multirow}
\usepackage{rotating}
\usepackage{cite}
\usepackage{threeparttable}
\usepackage{hyperref}

\begin{document}
\title{EC-SLAM: Effectively Constrained Neural RGB-D SLAM with Sparse TSDF Encoding and Global Bundle Adjustment}
\author{Guanghao Li\textsuperscript{\dag}\hspace{2em}Qi Chen\textsuperscript{\dag}\hspace{2em}Yuxiang Yan\hspace{2em}Jian Pu\textsuperscript{*}

\thanks{\dag\hspace{1pt}Indicates equal contribution.}

\thanks{*\hspace{1pt}Corresponding author.}

\thanks{G. Li, Y. Yan, and J. Pu are with the Institute of Science and Technology for Brain-Inspired
Intelligence, Fudan University, Shanghai, 200433, China (E-mails: \{ghli22, yxyan22\}@m.fudan.edu.cn, jianpu@fudan.edu.cn).}

\thanks{Q. Chen is with Shanghai Key Lab of Intelligent Information Processing and School of Computer Science, Fudan University, Shanghai 200433, China (E-mail: qichen21@m.fudan.edu.cn).}

}

\markboth{Journal of \LaTeX\ Class Files,~Vol.~18, No.~9, September~2020}%
{How to Use the IEEEtran \LaTeX \ Templates}

\maketitle

\begin{abstract}
We introduce EC-SLAM, a real-time dense RGB-D simultaneous localization and mapping (SLAM) system leveraging Neural Radiance Fields (NeRF). While recent NeRF-based SLAM systems have shown promising results, they have yet to fully exploit NeRF's potential to constrain pose optimization. EC-SLAM addresses this by using sparse parametric encodings and Truncated Signed Distance Fields (TSDF) to represent the map, enabling efficient fusion, reducing model parameters, and accelerating convergence. Our system also employs a globally constrained Bundle Adjustment (BA) strategy that capitalizes on NeRF's implicit loop closure correction capability, improving tracking accuracy by reinforcing constraints on keyframes most relevant to the current optimized frame. Furthermore, by integrating a feature-based and uniform sampling strategy that minimizes ineffective constraint points for pose optimization, we reduce the impact of random sampling in NeRF. Extensive evaluations on the Replica, ScanNet, and TUM datasets demonstrate state-of-the-art performance, with precise tracking and reconstruction accuracy achieved alongside real-time operation at up to 21 Hz. The source code is available at \href{https://github.com/Lightingooo/EC-SLAM}{https://github.com/Lightingooo/EC-SLAM}.

\end{abstract}
\begin{IEEEkeywords}
Neural Rendering,  Neural Radiance Field, Simultaneous Localization and Mapping.
\end{IEEEkeywords}
\begin{figure*}[!t]
    \centering
    \includegraphics[width=\linewidth]{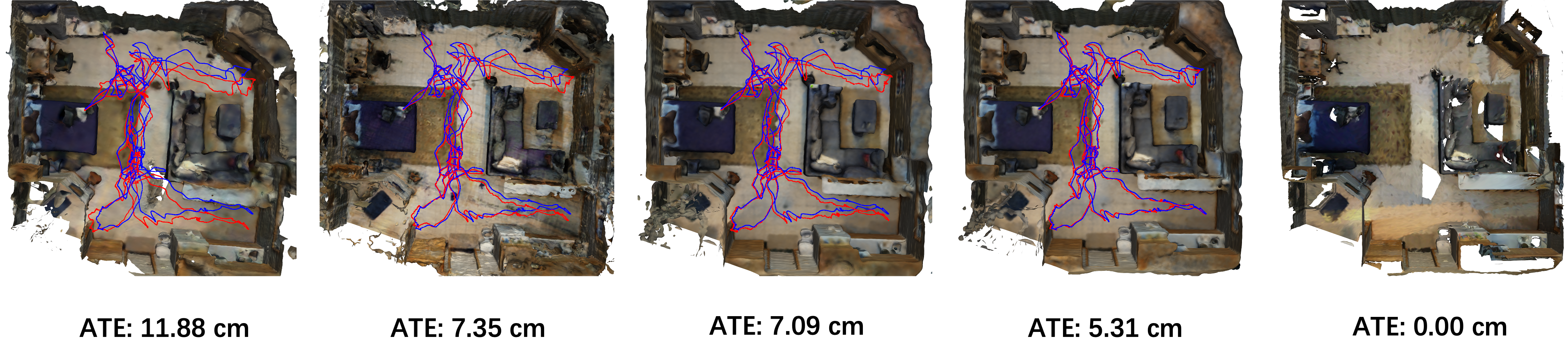}
    \begin{minipage}{\linewidth}
    \raggedright
    \textbf{\hspace{1.3em} NICE-SLAM \cite{zhu2022nice}}  \hspace{3.2em} \textbf{ESLAM \cite{johari2023eslam}} \hspace{4.5em}  \textbf{Co-SLAM \cite{wang2023co}} \hspace{5.5em}   \textbf{Ours}  \hspace{5.7em}   \textbf{Ground Truth}
    \end{minipage}
    \caption{3D reconstruction and tracking of the trajectory for scene0000 from the ScanNet dataset~\cite{dai2017scannet}. The blue line depicts the ground truth camera trajectory, while the red line shows the estimated camera trajectory. Our method demonstrates superior performance in tracking and quality of scene reconstruction compared to other RGB-D methods such as NICE-SLAM~\cite{zhu2022nice}, Co-SLAM~\cite{wang2023co}, and ESLAM~\cite{johari2023eslam}.}
    \label{fig:firstG}
\end{figure*}

\section{Introduction}
\label{sec:intro}
\IEEEPARstart{O}ver the past several decades, the field of real-time visual simultaneous localization and mapping (VSLAM) has rapidly advanced to become a cornerstone of research and development, particularly within the realms of autonomous driving and virtual reality. This groundbreaking technology, which enables the precise mapping of environments while simultaneously determining the observer's location within that environment, has been instrumental in pushing the boundaries of what's possible in navigation and immersive experiences. However, despite the impressive achievements of traditional VSLAM methodologies in terms of localization accuracy—allowing devices to pinpoint their position with remarkable precision—they often fall short when it comes to creating maps that are both dense and continuous. Such comprehensive and detailed mapping is crucial for the functionality of advanced applications, necessitating ongoing research and innovation to bridge this gap and meet the increasingly complex demands of these cutting-edge technologies.

Learning-based VSLAM systems have increasingly become a pivotal solution within the VSLAM domain, mainly due to their capacity for dense scene representation. Notably, VSLAM systems that utilize NeRF~\cite{mildenhall2020nerf} and its variants inputting RGB-D images have excelled in map and pose optimization via differentiable volume rendering. These systems are primarily categorized into two types based on their prediction methodologies: Multilayer Perceptron (MLP) based VSLAM \cite{sucar2021imap, kong2023vmap} and feature-decoder based VSLAM \cite{zhu2022nice, johari2023eslam, wang2023co}. While MLP based VSLAM systems employ one or more MLPs for scene representation, offering smooth depiction, they face a challenge in pose constraints due to slow convergence. This slow convergence is attributed to the necessity of training all parameters in each iteration. Additionally, as the training scenarios increase in size, this integrated neural network struggles to address the issue of catastrophic forgetting.

The feature-decoder based VSLAM systems avoid training all parameters within one iteration and simultaneously ensure the smoothness of the scene through handcrafted loss functions, thus relatively speeding up the training process. However, compared to traditional VSLAM methods, they still operate at a slower speed and have inferior tracking accuracy. Furthermore, due to speed and memory constraints, these systems rely on random sampling methods for joint local/global BA and the continuous tracking of new keyframes. Specifically, Co-SLAM \cite{wang2023co} selects a random number and sequence of keyframes and their corresponding pixels during global BA and randomly selects pixels during tracking. ESLAM \cite{johari2023eslam} randomly samples pixel positions during local BA and tracking. This randomness, in turn, affects the constraints on the poses that need to be optimized, thereby influencing the accuracy of tracking and reconstruction.

We introduce EC-SLAM, a learning-based dense RGB-D SLAM system to address the above challenges. We employ sparse parametric encodings with TSDF and implement strategies such as effectively constrained global BA, robust sampling, and an accelerating algorithm. This enables us to achieve precise mapping and tracking precision, as demonstrated in Figure \ref{fig:firstG}. Furthermore, we observed that SLAM systems based on NeRF inherently exhibit a specific skill for detecting loop closures. Our system significantly improves the capability of NeRF loop correction. NeRF's implicit loop detection is more natural and effective compared to the explicit loop detection methods used in classical SLAMs, such as descriptor and bag-of-words matching. Extensive evaluations and ablation analyses carried out on several datasets (Replica \cite{straub2019replica}, ScanNet \cite{dai2017scannet}, Tum \cite{sturm2012benchmark}) demonstrate that our system provides superior reconstruction and tracking accuracy compared to other cutting-edge systems at a speed of up to 21 Hz. 

The contributions of this paper can be summarized as follows. Firstly, we integrate sparse parametric encodings with TSDF in order to accurately represent and optimize maps. This method allows for the quick and precise reconstruction of relevant maps. Our second contribution is the development of a novel NeRF-based architecture that effectively enforces constraints within the global bundle adjustment framework. This architecture also takes advantage of the implicit loop closure capabilities of NeRF, which are manifested in the selection of sliding windows, the organization of keyframes, and the optimization of pixel sampling. This strategy greatly improves the accuracy and reliability of tracking and mapping. At last, we have created a reliable and immediate RGB-D dense SLAM system that takes advantage of the individual advantages of NeRF and classic SLAM systems for mapping and tracking. Our system has been proven effective through extensive experiments. Our strategy can be used as a robust benchmark for future works.

\section{Related Work}
\label{sec:review}

\noindent
\textbf{Traditional Dense VSLAM.} Compared to SLAM systems that perform sparse reconstruction \cite{davison2007monoslam, vincke2010design, vincke2012efficient, klein2007parallel, jimenez2017intelligible, forster2014svo, forster2016svo, mur2015orb, mur2017orb, campos2021orb, engel2017direct, mourikis2007multi, leutenegger2015keyframe, bloesch2015robust, mur2017visual, qin2018vins, von2018direct}, fewer systems execute dense reconstruction, which are more computationally intensive. However, they provide a detailed and accurate map. DTAM \cite{newcombe2011dtam}, and Mobilefusion \cite{ondruvska2015mobilefusion} were seminal direct methods using photometric error minimization. \cite{engel2014lsd, boikos2016semi, boikos2017high} were direct methods offering semi-dense reconstruction. Dense mapping SLAM systems also incorporated inertial measurements or depth data instead of visual-only input. KinectFusion \cite{newcombe2011kinectfusion} was a real-time RGB-D SLAM algorithm for 3D reconstruction and surface mapping, with limitations in handling drift over time. \cite{salas2013slam++, kerl2013dense, endres20133} employed loop closure detection to achieve better tracking and mapping results. Additionally, recent learning-based methods \cite{tateno2017cnn, teed2021droid, yang2020d3vo, li2018undeepvo, li2020deepslam, kang2019df, ummenhofer2017demon, zhou2018deeptam, bloesch2018codeslam, zhi2019scenecode, sucar2020nodeslam, teed2020deepv2d, tang2018ba, czarnowski2020deepfactors} have emerged. These methods can perform pose and depth estimation via a deep neural network. However, these emerging systems' scene representation and pipeline are still based on traditional systems.

\vspace{\baselineskip}
\noindent
\textbf{NeRF-based Dense VSLAM.} Although NeRF-based SLAM has lower tracking accuracy compared to traditional VSLAM, it is capable of real-time dense mapping and leverages this capability to utilize its implicit loop closure ability. It can mainly be divided into two broad categories: one integrates NeRF \cite{mildenhall2020nerf} with components from traditional SLAM systems \cite{chung2023orbeez, rosinol2022nerf, zhang2023go}, while the other solely rely on NeRF \cite{sucar2021imap, zhu2022nice, wang2023co, johari2023eslam, yang2022vox, zhu2023nicer, li2023dense, plgSLAM, loopyliso,SLAIMCartillier}. 

In the former type, NeRF-SLAM \cite{rosinol2022nerf} combined the tracking component of DROID-SLAM \cite{teed2021droid} with its NeRF-based networks, while Orbeez-SLAM \cite{chung2023orbeez} integrated with modules from ORB-SLAM3 \cite{campos2021orb}. GO-SLAM \cite{zhang2023go} extended the tracking component of DROID-SLAM \cite{teed2021droid} by several key features. In the latter type, iMAP \cite{sucar2021imap}, and NICE-SLAM \cite{zhu2022nice} used a single MLP and a coarse-to-fine feature grid to represent maps, respectively, but they suffered from slow convergence and struggled with scene details. Nicer-SLAM \cite{zhu2023nicer} performed mapping and tracking with monocular input instead of RGB-D input. For other RGB-D systems, Co-SLAM \cite{wang2023co} and ESLAM \cite{johari2023eslam} employed InstantNGP \cite{muller2022instant} and Tri-plane \cite{chan2022efficient} as their scene representations, respectively, both utilizing TSDF prediction for quicker convergence. In addition to the systems mentioned above, several SLAM systems~\cite{kong2023vmap,SNISLAMZHU,NIS-SLAMZhai} have integrated semantic information, which enriches the generated maps but also increases resource consumption. Furthermore, with the rise of 3D Gaussian Splatting (3DGS) ~\cite{3dgsKerbl}, many SLAM systems~\cite{matsuki2024gaussian, keetha2024splatam} have extended 3DGS into the SLAM domain, leading to high-quality rendering. However, current 3DGS-based systems still struggle with accurate surface reconstruction and tend to run slowly, falling short of the real-time requirements expected of SLAM systems.

Our system belongs to the type that solely relies on NeRF \cite{mildenhall2020nerf}. Unlike the systems mentioned above, we have meticulously optimized the number and sequence of keyframes participating in BA, as well as the method of pixel sampling, to implement an effectively constrained global BA and utilize NeRF \cite{mildenhall2020nerf} for implicit loop closure correction. Moreover,  our sparse parametric encodings and specialized loss functions contribute to the effect of real-time, small-parameter, and high-precision in mapping and tracking. This focus on optimization and precision underscores our system's advanced capabilities in the evolving landscape of VSLAM technology.
\section{Method}
\label{sec:Method}
    \begin{figure*}[!t]
        \centering
        \includegraphics[width=\linewidth]{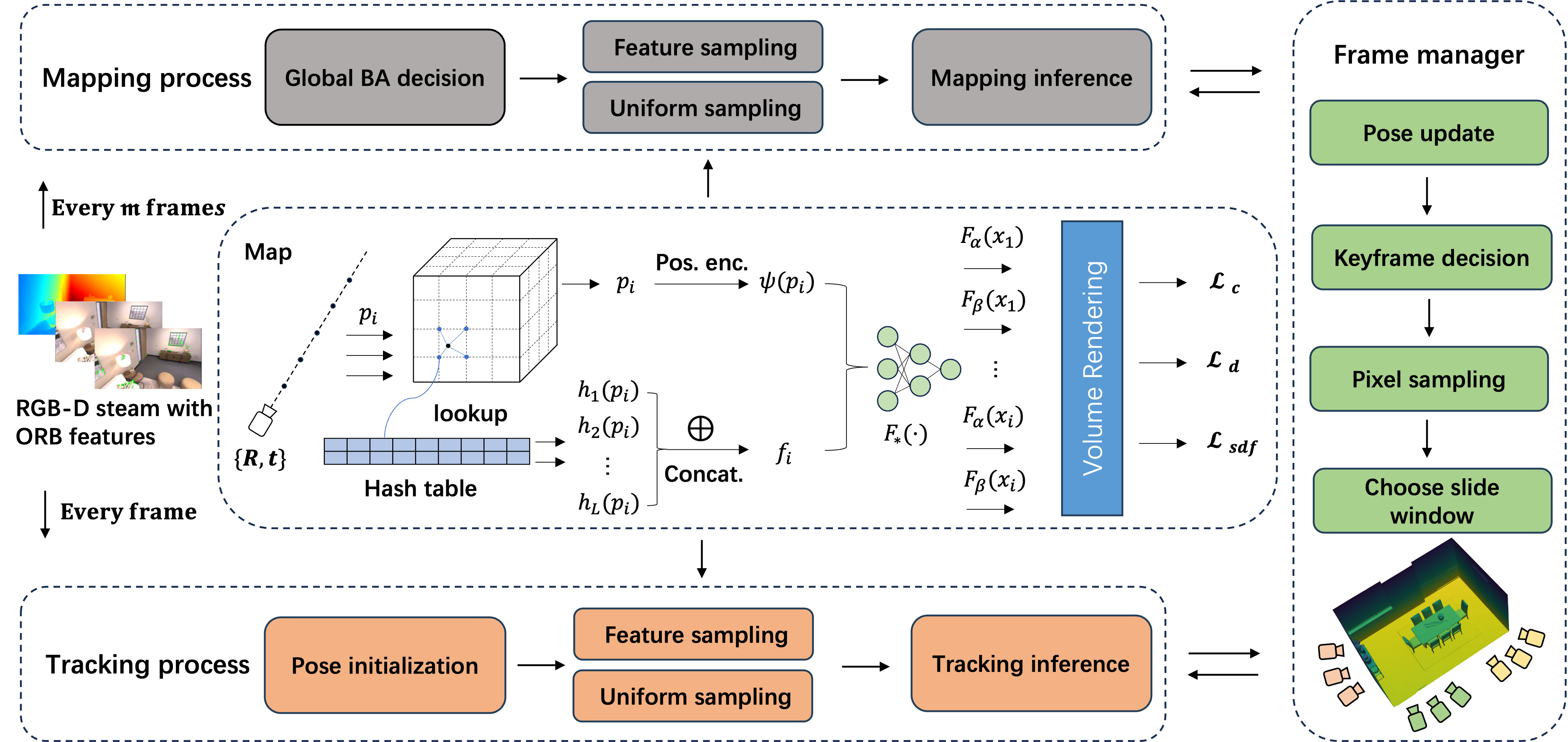} 
        \hfill
        \caption{The overview of EC-SLAM. (1) Mapping process: We jointly optimize the map and the poses of some keyframes in the sliding window. (2) Tracking process: We use a constant velocity motion model to initialize the tracking frame's pose, followed by iterative optimization. (3) Map: We employ a multi-resolution hash grid to store feature values for each point in space and use two MLPs to decode these features, obtaining color and TSDF values. Here $ \boldsymbol{f}_i $, $ \psi(\boldsymbol{p}_i) $ and $ F_*(\cdot)$ denotes the multi-resolution feature, position encoding and decoder function for a certain point, respectively. We compute color, depth, and TSDF losses to optimize the map and pose. $ \mathcal{L}_{c} $, $ \mathcal{L}_{d} $, and $ \mathcal{L}_{s} $ are the color loss, depth loss, and TSDF loss, respectively.}
        \label{fig:overview}
    \end{figure*}

     Given a set of sequential RGB-D frames $ \{ I_i, D_i \}_{i=1}^{M} $ with known camera intrinsics $ K \in \mathbb{R}^{3 \times 3} $, we predicts camera poses $ \{ R_i|t_i \}_{i=1}^{M} $ and an implicit TSDF map representation. As shown in Fig. \ref{fig:overview}, we established two threads to address this problem: the mapping and tracking threads. The mapping thread is responsible for jointly optimizing the map and the poses of the keyframes in the sliding window, while the tracking thread estimates the pose of the new image frame. Consistent with the previous system \cite{wang2023co}, we optimize the map after a fixed number of frames. The map, represented as a hash map, is a shared variable transmitted between different threads. Each thread performs volume rendering based on the corresponding pose and hash map, then computes the loss against the actual image. 
     
\subsection{Map Representation and Volume Rendering}
\label{subM:mapRep}
    \textbf{Map Representation.} The quantity of grid parameters can significantly influence the convergence speed and reconstruction accuracy when restricting the convergence speed to feature-decoder based methods. To mitigate this issue, we adopt the multi-resolution hash encoding proposed in \cite{muller2022instant} as our scene representation structure. For a point $ \boldsymbol{x}_i $ in space, we have:
    \begin{equation}
        \boldsymbol{f}_i = \bigoplus_{l=1}^{L} h_l(\boldsymbol{x}_i) ,
        \label{eq:hashFunc}
    \end{equation}
    where $ L $ denotes the number of resolution levels in the hash grid. $ h_l(\cdot) $ is the hash lookup and interpolation function that performs linear interpolation in the corresponding level for a certain point. $\boldsymbol{f}_i$ is the final feature obtained after concatenating each level's feature.
    
    In addition to the network architecture, we utilize the One-blob encoding~\cite{wang2023co, muller2019neural} for positional representation. We ascertain a given spatial point's One-blob encoding and hash features, which are subsequently fed into the corresponding TSDF and color decoders. Unlike \cite{wang2023co}, we do not employ the intermediate embedding generated by the TSDF decoder when decoding color. Thus, the inputs to both the TSDF and color decoders are identical. We observed that such an approach facilitates more rapid network convergence.

    \vspace{\baselineskip}
    \noindent
    \textbf{Volume Rendering.} We employ a uniform sampling strategy based on depth and spatial distribution to acquire points from the optical center to the object surface. In conjunction with the TSDF representation, we perform uniform sampling along the spatial ray and on the surface of the object, resulting in a total of $N=N_u+N_d$ sampled points in the space. Here, $ N_u $ and $ N_d $ represent the number of points sampled uniformly in space and based on depth. For depth-based sampling, we sample $ N_d $ points within the interval $ [d-d_t, d+d_t] $, where $ d $ is the object surface depth and $ d_t $ is a hyperparameter that defines the sampling density.
    
    As described in \cite{johari2023eslam, wang2023co}, predicting the TSDF values for spatial points instead of directly predicting their volumetric density can significantly enhance the speed and accuracy of network convergence because it allows constraints to be applied to the TSDF values at each point along a ray rather than constraining the entire ray. As detailed in Equ. \ref{eq:sdf2alpha}, we employ the method outlined in \cite{or2022stylesdf} to transform the predicted TSDF values into volumetric density:
    
    \begin{equation}
        \sigma\left(s_i\right)=\beta \cdot \operatorname{Sigmoid}\left(-\beta \cdot s_i\right),
        \label{eq:sdf2alpha}
    \end{equation}
    where $ s_i $ is the TSDF of a certain point, and $ \beta $ is a learnable parameter that controls the sharpness of the surface boundary \cite{johari2023eslam}. 

    After determining the volumetric density of all points along a ray, we follow the standard volume rendering procedure in \cite{mildenhall2020nerf} to render the weights for all points on the ray, thereby obtaining the final color and depth corresponding to the pixel associated with that ray (Equ. \ref{eq:renderFunc}). Here, $ N=N_u+N_d $ represents the total number of points sampled along the ray, $ w_n $ is the weight for space point $ n $, $ z_n $ is the sampled depth along each ray, and $ \hat{\boldsymbol{c}} $ and $ \hat{d} $ are the estimated color and depth.
    \begin{equation}
        \begin{gathered}
            w_n=\exp \left(-\sum_{i=1}^{n-1} \sigma\left(s_i\right)\right)\left(1-\exp \left(-\sigma\left(s_n\right)\right)\right), \\
            \hat{\boldsymbol{c}}=\sum_{n=1}^N w_n \boldsymbol{c}_n \quad \text { and } \quad \hat{d}=\sum_{n=1}^N w_n z_n.
        \end{gathered}
        \label{eq:renderFunc}
    \end{equation}

\subsection{Robust multi-threaded Pipeline}
\label{subM:MandT}
    \textbf{System Pipeline.} We introduce our system from two aspects: the mapping thread and the tracking thread. Similar to the majority of NeRF-SLAM systems \cite{sucar2021imap, zhu2022nice, wang2023co, johari2023eslam}, the mapping thread is responsible for both map initialization and subsequent mapping tasks. We randomly initialize the network parameters and fix the pose of the first image frame for mapping. In the subsequent iterations, we perform a mapping operation at fixed intervals of image frames. In each iteration, we first decide on the sliding window containing keyframes, then perform our robust pixel sampling method on all the keyframes in the sliding window. Finally, we jointly optimize our map and the poses of some keyframes in the sliding window, which is called effectively constrained global BA.
    
    During tracking, we first calculate the initial pose of the frame to be tracked based on a constant velocity motion model. Then, we sample the tracking image frame using a sampling strategy similar to the mapping thread. Finally, we fix our map parameters and iteratively optimize the input pose. 

    \vspace{\baselineskip}
    \noindent
    \textbf{Effectively Constrained Global BA.} In our proposed methodology, we address the limitations observed in certain NeRF-based SLAM systems. Previous systems \cite{zhu2022nice, johari2023eslam} often employ a process where pixels and points sampled from the current mapping frame are projected across all other keyframes to determine the sliding window. This process, prone to randomness, proves to be excessively time-consuming. Furthermore, these systems typically utilize a sliding window of limited size, which does not adequately support global optimization or facilitate implicit loop closure correction. On the other hand, some systems \cite{wang2023co} treat the sliding window as an approximation of all keyframes. Although this considers global optimization, it does not identify the frames most correlated with the current frame's pose. Furthermore, as the number of keyframes grows, the number of sampled pixels per keyframe decreases, and the distribution of sampled pixels among frames becomes uneven, consequently weakening the pixel constraints on optimizing keyframe poses.
    
    We adopt a simple, fast ,and effective strategy to determine the number and order of keyframes in the sliding window. To eliminate the impact of randomness, we use only the pose to decide the sliding window. Precisely, we first calculate the optical center distance $d$ and parallax angle $r$ between all keyframes and the current frame's camera, then filter out keyframes with large parallax angles and sort them based on distance. 
    
    Regarding the selection of keyframe quantity, to ensure the effectiveness of global BA, we set a maximum number threshold based on the number of sampled pixels. AS long as the number of keyframes in the sliding window does not exceed this threshold, we strive to include keyframes that can form loop closure constraints with the current frame, ensuring the presence of historical and the most recent frames. This approach not only forms practical global BA constraints but also fully leverages the inherent implicit loop closure capabilities of NeRF \cite{mildenhall2020nerf}.

    \vspace{\baselineskip}
    \noindent
    \textbf{Robust Pixel Sampling Method.} A distinctive feature of NeRF-based SLAM \cite{mildenhall2020nerf} is random pixel sampling, which significantly influences the results. We employ two sampling methods that mitigate this issue: feature point sampling and random uniform sampling. We initially extract feature points \cite{shi1994good} for each image to serve as a subset of the sampled pixel points. For the remaining pixels to be sampled, we establish a grid over the image and perform uniform sampling within this grid.

    Moreover, sampling at each iteration was excessively time-consuming, so we perform a one-time sampling in the camera coordinate system before iteration. During each iteration, we project the relevant pixels to the world coordinate system by using the current optimized pose. Therefore, our robust pixel sampling method significantly reduces time consumption on iterative sampling without substantially increasing memory usage.

\subsection{Objective Functions}
\label{subM:loss}
    The objective function for measuring network performance consists of three components: color loss, depth loss, and TSDF loss. We employ the L2 loss to quantify the discrepancy between rendered and actual values for color and depth:
    \begin{equation}
    \begin{gathered}
        \mathcal{L}_d=\frac{\lambda_{ud}}{|R_u|} \sum_{r \in R_u}(\hat{d}_r-d_r)^2 + \frac{\lambda_{fd}}{|R_f|} \sum_{r \in R_f}(\hat{d}_r-d_r)^2, \\
        \mathcal{L}_c=\frac{\lambda_{uc}}{|R_u|} \sum_{r \in R_u}(\hat{\boldsymbol{c}}_r-\boldsymbol{c}_r)^2 + \frac{\lambda_{fc}}{|R_f|} \sum_{r \in R_f}(\hat{\boldsymbol{c}}_r-\boldsymbol{c}_r)^2.
        \label{eq:lossFcd}
    \end{gathered}
    \end{equation}
    $ R_f $ and $ R_u $ are the feature-based and uniform sampled rays, respectively. $R = R_f + R_u $ represents all the sampled rays. $ \lambda_{*}$ represents the weight for each loss.

    To further reinforce constraints on each sampled spatial point along each ray, we calculate the corresponding TSDF loss for each point based on its distance from the scene surface, following the approach in \cite{johari2023eslam}. For points beyond the truncation region $ T $, we assess the difference between the predicted TSDF value and the actual value of 1:
    \begin{equation}
    \mathcal{L}_{f s}=\frac{1}{|R|} \sum_{r \in R} \frac{1}{\left|P_r^{f s}\right|} \sum_{p \in P_r^{f s}}\left(s_{p}-1\right)^2,
    \label{eq:lossFs}
    \end{equation}
    where $ P_r^{f s} $ denotes the points along the ray beyond the truncation region $ T $.  
    For points within the truncation region $ T $, the closer a point is to the surface, the more its TSDF value approaches zero, and TSDF values are negative for points inside the surface:
    \begin{equation}
    \begin{gathered}
        \begin{aligned}
            & \mathcal{L}_t= 
            & \frac{1}{|R|} \sum_{r \in R} \frac{1}{\left|P_r^t\right|} \sum_{p \in P_r^t}\left(z_p+s_{p} \cdot T-d_r\right)^2,
        \end{aligned}
        \\
        \begin{aligned}
            & \mathcal{L}_m= 
            & \frac{1}{|R|} \sum_{r \in R} \frac{1}{\left|P_r^m\right|} \sum_{p \in P_r^m}\left(z_p+s_{p} \cdot T-d_r\right)^2,
        \end{aligned}
        \label{eq:lossFt}
    \end{gathered}
    \end{equation}
    here, $ P_r^{m} $ denotes the set of points that $ \left| z_p -  d_r \right| < \gamma T $ for a ray $ r $, and $ P_r^{t} $ is the other points in the truncation region. $ \gamma $ is the parameter that controls the number of spatial points within $ P_r^m $.
    
    The total loss is the weighted sum of the losses above, where the weights act as hyperparameters reflecting each loss component's relative importance:
    \begin{equation}
    \mathcal{L}=\lambda_{f s} \mathcal{L}_{f s}+\lambda_{m} \mathcal{L}_{m}+\lambda_{t} \mathcal{L}_{t}+ \mathcal{L}_d+ \mathcal{L}_c.
    \label{eq:lossTotal}
    \end{equation}
    We apply smooth loss from \cite{wang2023co} at the last image frame to make the scene more consistent.

\begin{figure*}[!t]
    \centering
    {\raggedleft room-0 \hspace{13em} room-2  \hspace{12em} office-3 \hspace{7em} \par }
    \vspace{0.5em}
    \begin{minipage}[t]{0.02\textwidth}
    \begin{sideways}
        \hspace{2em} GT \hspace{3.8em} Ours \hspace{2.6em} Co-S \cite{wang2023co} \hspace{2.1em} E-S \cite{johari2023eslam} \hspace{1.7em} NICE-S \cite{zhu2022nice}
    \end{sideways}
    \end{minipage}%
    \begin{minipage}[t]{0.98\textwidth}
      \centering
      \includegraphics[width=\linewidth]{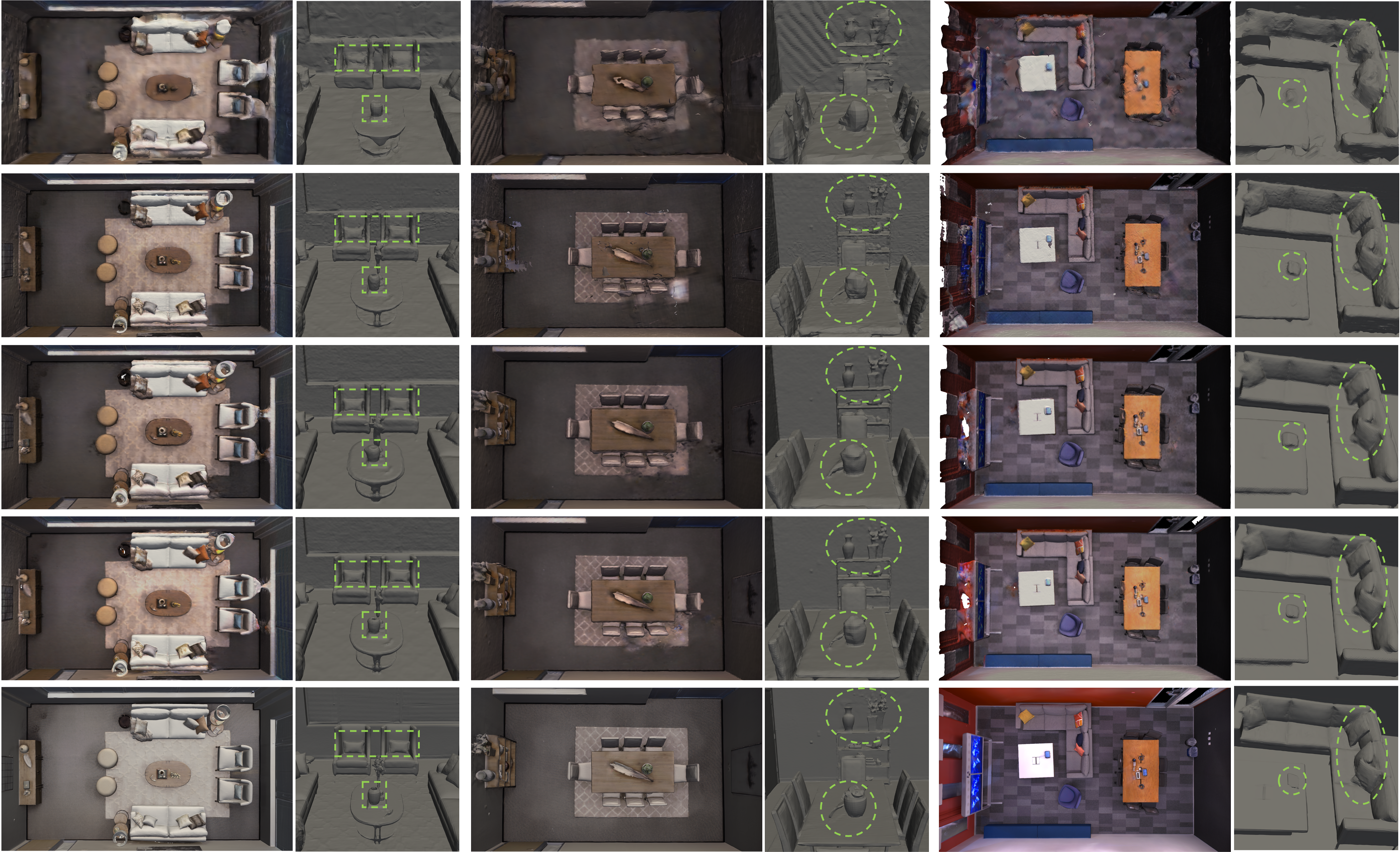} 
    \end{minipage}%
    \hfill
    \caption{The reconstruction results of our system with other SOTA NeRF-based RGB-D dense visual SLAM systems on the Replica dataset \cite{straub2019replica}. Our system reconstructed more accurate scenes compared with other systems, and the reconstruction details encircled by the green dashed line further demonstrate our system's better trade-off between sharpness and smoothness.}
    \label{fig:replica}
\end{figure*}

\begin{table*}[htbp]
\begin{center}
    \caption{Mapping and Tracking Results on the Replica dataset \cite{straub2019replica}}
    \label{tab:replicaDetail}%
    \renewcommand\arraystretch{0.8}
    \tabcolsep=0.33cm
    \begin{tabular}{c|c|cccc|cc|c}
    \toprule
          & \multirow{2}[2]{*}{Methods} & \multicolumn{4}{c|}{Reconstruction(cm)} & \multicolumn{2}{c|}{Localization(cm)} \\
          &       & Depth L1$\downarrow$ & Acc.$\downarrow$  & Comp.$\downarrow$ & Comp. Ratio(\%)$\uparrow$ & ATE RMSE$\downarrow$ & ATE Mean$\downarrow$ & FPS \\
    \midrule
    \multirow{4}[2]{*}{\begin{sideways}Room0\end{sideways}} 
          & iMAP \cite{sucar2021imap} & 5.17 ± 0.33  & 4.14 ± 0.25  & 5.99 ± 0.22 & 77.84 ± 1.12 & 5.43 ± 1.28  & 3.20 ± 0.75 & 0.4 \\
          & NICE-SLAM \cite{zhu2022nice} & 1.83 ± 0.13 & 2.48 ± 0.06 & 2.68 ± 0.08 & 91.66 ± 0.13 & 1.64 ± 0.11  & 1.43 ± 0.09 & 0.9 \\
          & Co-SLAM \cite{wang2023co} & 1.05 ± 0.03 & 2.03 ± 0.05 & 2.07 ± 0.04 & 95.16 ± 0.04 & 0.70 ± 0.03 & 0.63 ± 0.02 & 17.1 \\
          & ESLAM \cite{johari2023eslam} & 0.86 ± 0.03  & 2.52 ± 0.04 & \textbf{1.98 ± 0.01} & \textbf{96.04 ± 0.05} & 0.70 ± 0.13 & 0.60 ± 0.06 & 12.1 \\
          & Ours (Lite) & 0.66 ± 0.01  & 1.90 ± 0.04 & 2.09 ± 0.04 & 95.36 ± 0.06 & 0.77 ± 0.03  & 0.66 ± 0.03  & \textbf{21.4} \\
          & Ours (Full) & \textbf{0.59 ± 0.01} & \textbf{1.89 ± 0.02}  & 2.07 ± 0.03 & 95.79 ± 0.04  & \textbf{0.27 ± 0.01}  & \textbf{0.24 ± 0.01} & 11.9 \\
    \midrule
    \multirow{4}[2]{*}{\begin{sideways}Room1\end{sideways}} 
          & iMAP \cite{sucar2021imap}& 3.56 ± 0.68  & 3.16 ± 0.36 & 4.57 ± 0.40 & 85.39 ± 1.05 & 3.22 ± 0.36 & 2.66 ± 0.26 & 0.4 \\
          & NICE-SLAM \cite{zhu2022nice}& 1.41 ± 0.16 & 2.14 ± 0.06 & 2.23 ± 0.09 & 93.42 ± 0.19 & 2.08 ± 0.08 & 1.70 ± 0.29 & 0.9 \\
          & Co-SLAM \cite{wang2023co}& \textbf{0.85 ± 0.02} & 1.60 ± 0.04 & 1.86 ± 0.05 & 95.19 ± 0.03 & 1.09 ± 0.34 & 0.96 ± 0.35 & 17.1 \\
          & ESLAM \cite{johari2023eslam}& 0.88 ± 0.04  & 2.51 ± 0.05   & 1.78 ± 0.01  & 95.09 ± 0.06& 0.70 ± 0.02 & 0.55 ± 0.02 & 12.1 \\
          & Ours (Lite) & 0.86 ± 0.01 & 1.67 ± 0.04 & 1.77 ± 0.05 & 95.49 ± 0.07 & 1.07 ± 0.16 & 0.95 ± 0.16 & \textbf{21.4} \\
          & Ours (Full) & 0.89 ± 0.01 & \textbf{1.56 ± 0.02} & \textbf{1.66 ± 0.05} & \textbf{96.51 ± 0.05} & \textbf{0.28 ± 0.02}  & \textbf{0.24 ± 0.02} & 11.9 \\
    \midrule
    \multirow{4}[2]{*}{\begin{sideways}Room2\end{sideways}} 
          & iMAP \cite{sucar2021imap}& 5.87 ± 0.78  & 3.96 ± 0.39 & 5.23 ± 0.27 & 79.02 ± 1.63 & 2.85 ± 0.16 & 2.41 ± 0.20 & 0.4 \\
          & NICE-SLAM \cite{zhu2022nice}& 2.22 ± 0.20 & 2.21 ± 0.09 & 2.85 ± 0.08 & 91.32 ± 0.29 & 1.80 ± 0.28 & 1.41 ± 0.24 & 0.9 \\
          & Co-SLAM \cite{wang2023co}& 2.37 ± 0.02 & 1.95 ± 0.04 & 1.99 ± 0.05 & 93.48 ± 0.04 & 1.21 ± 0.14 & 0.82 ± 0.02 & 17.1 \\
          & ESLAM \cite{johari2023eslam}& 1.18 ± 0.03 & 1.76 ± 0.06 & 1.78 ± 0.01 & 95.88 ± 0.06 & 0.51 ± 0.01 & 0.42 ± 0.01 & 12.1 \\
          & Ours (Lite) & \textbf{1.10 ± 0.01} & 1.72 ± 0.03 & 1.67 ± 0.05 & 95.95 ± 0.06 & 0.64 ± 0.01  & 0.55 ± 0.02  & \textbf{21.4} \\
          & Ours (Full) & 1.11 ± 0.01 & \textbf{1.69 ± 0.01} & \textbf{1.59 ± 0.04} & \textbf{97.30 ± 0.04} & \textbf{0.32 ± 0.03}  & \textbf{0.29 ± 0.02}  & 11.9 \\
    \midrule
    \multirow{4}[1]{*}{\begin{sideways}Office0\end{sideways}} 
          & iMAP \cite{sucar2021imap}& 3.99 ± 0.34  & 3.37 ± 0.32 & 3.91 ± 0.37 & 83.01 ± 1.60 & 2.60 ± 0.82 & 1.74 ± 0.90 & 0.4\\
          & NICE-SLAM \cite{zhu2022nice}& 1.45 ± 0.14 & 1.87 ± 0.06 & 1.92 ± 0.09 & 94.80 ± 0.39 & 1.23 ± 0.24 & 1.12 ± 0.22 & 0.9\\
          & Co-SLAM \cite{wang2023co}& 1.24 ± 0.02 & 1.47 ± 0.05 & 1.63 ± 0.05 & 96.09 ± 0.05 & 0.56 ± 0.01 & 0.49 ± 0.01 & 17.1\\
          & ESLAM \cite{johari2023eslam}& 0.77 ± 0.02 & 1.61 ± 0.07 & 1.50 ± 0.01 & 97.32 ± 0.05 & 0.56 ± 0.04 & 0.41 ± 0.03 & 12.1\\
          & Ours (Lite) & \textbf{0.71 ± 0.01} & 1.42 ± 0.04 & 1.37 ± 0.04 & 97.70 ± 0.07 & 0.58 ± 0.03 & 0.49 ± 0.02  & \textbf{21.4}\\
          & Ours (Full) & 0.72 ± 0.01 & \textbf{1.39 ± 0.03} & \textbf{1.33 ± 0.04} & \textbf{97.79 ± 0.05} & \textbf{0.24 ± 0.01}  & \textbf{0.21 ± 0.00}  & 11.9\\
    \midrule
    \multirow{4}[1]{*}{\begin{sideways}Office1\end{sideways}} 
          & iMAP \cite{sucar2021imap}& 3.81 ± 0.39 & 2.13 ± 0.25 & 3.97 ± 0.24 & 88.05 ± 1.75 & 1.30 ± 0.23 & 1.18 ± 0.13 & 0.4\\
          & NICE-SLAM \cite{zhu2022nice}& 1.64 ± 0.11 & 1.62 ± 0.13 & 1.86 ± 0.10 & 93.94 ± 0.21 & 0.79 ± 0.17 & 0.74 ± 0.19 & 0.9\\
          & Co-SLAM \cite{wang2023co}& 1.48 ± 0.02 & 1.27 ± 0.04 & 1.64 ± 0.04 & 94.55 ± 0.03 & 0.60 ± 0.06 & 0.53 ± 0.06 & 17.1\\
          & ESLAM \cite{johari2023eslam}& 1.22 ± 0.02 & 1.98 ± 0.08 & \textbf{1.41 ± 0.01} & 96.66 ± 0.08 & 0.54 ± 0.04 & 0.45 ± 0.05 & 12.1\\
          & Ours (Lite) & 1.25 ± 0.01 & 1.30 ± 0.03 & 1.53 ± 0.05 & 95.73 ± 0.07 & 0.70 ± 0.05  & 0.63 ± 0.05  & \textbf{21.4}\\
          & Ours (Full) & \textbf{1.15 ± 0.01} & \textbf{1.22 ± 0.02} & 1.46 ± 0.05 & \textbf{96.84 ± 0.05} & \textbf{0.25 ± 0.01}  & \textbf{0.23 ± 0.01}  & 11.9\\
    \midrule
    \multirow{4}[2]{*}{\begin{sideways}Office2\end{sideways}} 
          & iMAP \cite{sucar2021imap}& 4.02 ± 0.53 & 4.18 ± 0.41 & 4.89 ± 0.37 & 79.17 ± 1.16 & 5.94 ± 1.60 & 4.08 ± 0.78 & 0.4\\
          & NICE-SLAM \cite{zhu2022nice}& 2.71 ± 0.13 & 3.32 ± 0.13 & 3.23 ± 0.08 & 88.12 ± 0.28 & 1.69 ± 0.14 & 1.42 ± 0.10 & 0.9\\
          & Co-SLAM \cite{wang2023co}& 1.86 ± 0.07  & 2.74 ± 0.08 & 2.48 ± 0.07 & 91.63 ± 0.06 & 2.08 ± 0.02 & 1.84 ± 0.02 & 17.1\\
          & ESLAM \cite{johari2023eslam}& \textbf{1.06 ± 0.02} & 2.87 ± 0.09 & 2.05 ± 0.02 & 94.38 ± 0.02 & 0.57 ± 0.09 & 0.46 ± 0.03 & 12.1\\
          & Ours (Lite) & 1.44 ± 0.01 & 3.13 ± 0.04 & \textbf{1.99 ± 0.06} & 94.20 ± 0.06 & 0.69 ± 0.03  & 0.60 ± 0.03 & \textbf{21.4} \\
          & Ours (Full) & 1.36 ± 0.01 & \textbf{2.39 ± 0.02} & 2.00 ± 0.05 & \textbf{94.72 ± 0.05} & \textbf{0.31 ± 0.02}  & \textbf{0.29 ± 0.02}  & 11.9\\
    \midrule
    \multirow{4}[2]{*}{\begin{sideways}Office3\end{sideways}} 
          & iMAP \cite{sucar2021imap}& 5.71 ± 0.95 & 4.28 ± 0.33 & 5.65 ± 0.20 & 73.42 ± 0.65 & 5.18 ± 1.24 & 4.16 ± 0.78 & 0.4\\
          & NICE-SLAM \cite{zhu2022nice}& 2.17 ± 0.40 & 3.05 ± 0.20 & 3.27 ± 0.09 & 87.53 ± 0.45 & 3.90 ± 1.34 & 2.31 ± 0.31 & 0.9\\
          & Co-SLAM \cite{wang2023co}& 1.66 ± 0.03 & 3.01 ± 0.05 & 2.77 ± 0.04 & 90.62 ± 0.03 & 1.58 ± 0.04 & 1.50 ± 0.05 & 17.1\\
          & ESLAM \cite{johari2023eslam}&1.02 ± 0.04 & 2.53 ± 0.07 & \textbf{2.27 ± 0.01} & 94.01 ± 0.03 & 0.71 ± 0.02 & 0.60 ± 0.03 & 12.1\\
          & Ours (Lite) & 0.84 ± 0.01 & 2.60 ± 0.03 & 2.36 ± 0.05 & 93.36 ± 0.05 & 0.88 ± 0.01  & 0.74 ± 0.01  & \textbf{21.4}\\
          & Ours (Full) & \textbf{0.79 ± 0.01} & \textbf{2.49 ± 0.02} & 2.33 ± 0.04 & \textbf{94.13 ± 0.04} & \textbf{0.38 ± 0.03}  & \textbf{0.35 ± 0.03 } & 11.9\\
    \midrule
    \multirow{4}[2]{*}{\begin{sideways}Office4\end{sideways}} 
          & iMAP \cite{sucar2021imap}& 5.81 ± 1.06 & 4.53 ± 0.24 & 6.81 ± 0.20 & 74.29 ± 0.93 & 2.42 ± 0.23 & 2.04 ± 0.16 & 0.4\\
          & NICE-SLAM \cite{zhu2022nice}& 2.10 ± 0.41 & 2.58 ± 0.23 & 3.72 ± 0.24 & 87.08 ± 1.35 & 2.77 ± 0.60 & 2.22 ± 0.39 & 0.9\\
          & Co-SLAM \cite{wang2023co}& 1.54 ± 0.02 & 2.41 ± 0.06 & 2.50 ± 0.05 & 90.32 ± 0.04 & 0.71 ± 0.01 & 0.62 ± 0.02 & 17.1\\
          & ESLAM \cite{johari2023eslam}& 1.10 ± 0.04 & 2.14 ± 0.10 & \textbf{2.28 ± 0.01} & 93.10 ± 0.14 & 0.62 ± 0.03 & 0.51 ± 0.02 & 12.1\\
          & Ours (Lite) & 0.80 ± 0.01 & 2.15 ± 0.03 & 2.47 ± 0.06 & 91.95 ± 0.05 & 0.89 ± 0.07  & 0.78 ± 0.07  & \textbf{21.4}\\
          & Ours (Full) & \textbf{0.79 ± 0.01} & \textbf{1.96 ± 0.02} & 2.36 ± 0.05 & \textbf{93.24 ± 0.04} & \textbf{0.30 ± 0.03}  & \textbf{0.28 ± 0.03}  & 11.9\\
    \midrule
    \multirow{4}[2]{*}{\begin{sideways}Average\end{sideways}} 
          & iMAP \cite{sucar2021imap} & 4.74 ± 0.63  & 3.71 ± 0.31  & 5.12 ± 0.29  & 80.02 ± 1.23  & 3.61 ± 0.74 & 2.68 ± 0.49 & 0.4 \\
          & NICE-SLAM \cite{zhu2022nice} & 1.94 ± 0.21  & 2.41 ± 0.12  & 2.72 ± 0.11  & 90.98 ± 0.41  & 1.98 ± 0.37 & 1.54 ± 0.23 & 0.9 \\
          & Co-SLAM \cite{wang2023co} & 1.50 ± 0.03  & 2.06 ± 0.05 & 2.13 ± 0.05 & 93.38 ± 0.04  & 1.07 ± 0.08 & 0.93 ± 0.07  & 17.1 \\
          & ESLAM \cite{johari2023eslam} & 1.01 ± 0.03    & 2.24 ± 0.07  & 1.88 ± 0.08   &  95.31 ± 0.06  & 0.62 ± 0.05 & 0.51 ± 0.03 & 12.1 \\
          & Ours (Lite)  & 0.96 ± 0.01  & 1.99 ± 0.04  & 1.90 ± 0.05  & 94.96 ± 0.07  & 0.77 ± 0.05  & 0.67 ± 0.05 & \textbf{21.4} \\
          & Ours (Full)  & \textbf{0.93 ± 0.01} & \textbf{1.83 ± 0.02}  & \textbf{1.86 ± 0.05}  & \textbf{95.79 ± 0.04}  & \textbf{0.29 ± 0.02}  & \textbf{0.27 ± 0.02} & 11.9 \\
    \bottomrule
    \end{tabular}%
    \begin{tablenotes}
    \footnotesize
    Quantitative comparison of our system with other state-of-the-art NeRF-based RGB-D dense VSLAM systems on the scenes of the Replica dataset \cite{straub2019replica}. Numbers in \textbf{bold black font} represent the best results in each column. Our tracking results outperform those of other systems, and our mapping results reach state-of-the-art levels with the minor model parameters and highest FPS.
    \end{tablenotes}
\end{center}
\end{table*}%

\section{Experiments}

\label{sec:exp}
    This section demonstrates that our system outperforms current NeRF and 3DGS based systems on three widely-used datasets while maintaining a rapid processing speed. The ablation studies further confirm the efficacy of our approach.

\subsection{Experimental Setup}
    \noindent
    \textbf{Baselines.} We benchmark our method against seven existing state-of-the-art NeRF based RGB-D dense VSLAM methods: iMAP \cite{sucar2021imap}, NICE-SLAM \cite{zhu2022nice}, ESLAM \cite{johari2023eslam}, Co-SLAM \cite{wang2023co}, Loopy-SLAM~\cite{loopyliso}, SNI-SLAM~\cite{SNISLAMZHU}, NIS-SLAM~\cite{NIS-SLAMZhai}, two 3DGS-based methods: Gaussian Splatting SLAM~\cite{matsuki2024gaussian}, SplaTAM~\cite{keetha2024splatam} and three classic methods: Kintinuous \cite{newcombe2011kinectfusion}, BAD-SLAM \cite{schops2019bad}, ORB-SLAM2 \cite{mur2017orb}. 

    For NeRF-based methods, we note that Loopy-SLAM~\cite{loopyliso} is neural point based method, which can get high reconstruction results but operates at a slow speed. SNI-SLAM~\cite{SNISLAMZHU}, NIS-SLAM~\cite{NIS-SLAMZhai} are semantic NeRF based SLAM systems. Besides, 3DGS based methods are hard to extract mesh since the surface generated from these methods is ambiguous. For a fair comparison, we only compare the tracking results with these systems.

    \noindent
    \textbf{Datasets.} We evaluate our method on three standard 3D benchmarks. Following \cite{zhu2022nice, wang2023co, johari2023eslam}, we quantitatively evaluate the reconstruction and tracking quality on eight synthetic scenes (room0, room1, room2, office0, office1, office2, office3, office4) from Replica \cite{straub2019replica}. We also evaluate the tracking results on six scenes (scene0000, scene0059, scene0106, scene0169, scene0181, scene0207) from ScanNet \cite{dai2017scannet} and three scenes (fr1/desk, fr2/xyz, fr3/office) from TUM RGB-D \cite{sturm2012benchmark} datasets. 

    \noindent
    \textbf{Metrics.} Our assessment of reconstruction quality includes metrics such as Depth L1 (in cm), Accuracy (in cm), Completion (in cm), and Completion Ratio (in \%) with a threshold set at 5 cm. In line with \cite{wang2023co, zhu2022nice}, we exclude areas not observed by any camera frustum and apply mesh culling to remove extraneous points within the camera's view but outside the intended scene. We utilize the ATE RMSE metric \cite{sturm2012benchmark} (in cm for Replica \cite{straub2019replica}, m for others) for camera tracking evaluation. In addition, we calculate the Frames Per Second (FPS) following the way in Co-SLAM \cite{wang2023co}.

    \noindent
    \textbf{Implementation Details} We run our system on a desktop PC with an Intel Core i7-12700 CPU and NVIDIA RTX 3090 GPU. The resolution to generate the scene mesh is set to 0.02 m for all systems to compare the reconstruction metric fairly. The decoders used for predicting SDF and color consist of two two-layer fully connected networks whose hidden feature size is 32. Following the configuration of \cite{wang2023co}, we sample 5\% of the total pixels for each keyframe using our sampling strategy and store them in the keyframe set. The number of iterations during mapping and tracking and the number of points sampled on the rays corresponding to pixels varies across different datasets. We use two configurations for the Replica dataset \cite{straub2019replica}, which differs in the iteration times in mapping and tracking.
    
    In the mapping thread, we perform mapping every five frames. For the Replica dataset \cite{straub2019replica}, we employ two parameter configurations. In our lite version, the mapping thread iterates 10 times, sampling 2000 pixels per iteration. In our full version, it iterates 20 times, sampling 4000 pixels each time. The lite and full versions operate at speeds of 21 Hz and 12 Hz, respectively. For the ScanNet dataset \cite{dai2017scannet}, the mapping thread iterates 10 times with speed of 8 Hz, sampling 4000 pixels per iteration. For the TUM dataset \cite{sturm2012benchmark}, it iterates 20 times with speed of 7 Hz, sampling 2000 pixels each time.
    
    In the tracking thread, we track every frame and use a constant velocity assumption to compute the initial pose of the tracking frame. For the two versions of the Replica dataset \cite{straub2019replica}, the lite version iterates 8 times, sampling 1000 pixels per iteration, while the full version iterates 20 times, sampling 2000 pixels each time. In the ScanNet dataset \cite{dai2017scannet}, we iterate 10 times, sampling 2000 pixels per iteration. In the TUM dataset \cite{sturm2012benchmark}, we iterate 20 times, sampling 2000 pixels each time.
    
    For the number of samples per ray, following the approach of Co-SLAM \cite{wang2023co}, we employ both spatially uniform sampling and depth-guided sampling along a ray. We set $N_u$ to 32 and $N_d$ to 11 for the Replica dataset \cite{straub2019replica}. For ScanNet dataset \cite{dai2017scannet}, we set $N_u$ to 96 and $N_d$ to 21. On the TUM dataset \cite{sturm2012benchmark}, $N_u$ is set to 64 and $N_d$ to 21.
    
    Please note that we have implemented a series of acceleration optimizations, such as pre-sampling pixels in the camera coordinate system and transforming them into the world coordinate system, thereby significantly enhancing the efficiency of our system. These acceleration optimizations enable our system to operate up to 50\% faster than other systems under the same parameter configurations across various datasets and scenarios.
    
\subsection{Experimental Results}
    \noindent
    \textbf{Evaluation on Replica \cite{straub2019replica}.} We evaluated the reconstruction and tracking performance of our system in eight scenes of the Replica dataset \cite{straub2019replica}, with results detailed in Tab. \ref{tab:replicaDetail} and Tab.~\ref{tab:moretracking}. Each metric consists of the mean and variance obtained by randomly running the system five times on each scene of the dataset. It can be observed that other methods exhibit significant fluctuations in accuracy across different scenes, indicating their poor robustness. This is attributed to the randomness of sampling in NeRF. Both configurations of our system outperform other methods among most scenes with speed of 12-21 Hz. For the Lite configuration, we achieved a precision similar to ESLAM \cite{johari2023eslam} (the previous state-of-the-art) with an FPS of 21Hz. For the Full configuration, we improved the precision by 50\% relative to ESLAM \cite{johari2023eslam} , at a similar FPS of approximately 12Hz.
    
    Regarding mapping metrics, our system also reached SOTA levels, and our model has a minimal number of parameters (Tab.~\ref{tab:timeEfficiency}). The qualitative reconstruction results of the Replica dataset \cite{straub2019replica} are shown in Fig. \ref{fig:replica}, where our system constructs more accurate scenes than other systems. The mapping results of other systems are either overly smooth, failing to adequately display high-frequency information, or excessively sharp, unable to represent the natural continuity of the scene. The reconstructed scenes of our system exhibit a good tradeoff between sharpness and smoothness, which is particularly evident in the reconstruction of small objects within the scenes, benefiting from our accurate pose estimation and sampling strategy.

    \begin{table}[htbp]
        \centering
        \tabcolsep=0.1cm
        \caption{Tracking Results on 3DGS-based and NeRF-based Systems}
        \begin{tabular}{r|ccccccccc|c}
        \hline 
        \multirow{2}[2]{*}{Methods} & \multicolumn{9}{c|}{Replica} & \multirow{2}[2]{*}{FPS} \\
         & r0 & r1 & r2 & o0 & o1 & o2 & o3 & o4 & Avg. \\ 
        \hline
        \multicolumn{1}{c|}{SNI~\cite{SNISLAMZHU}} & 0.33 & 0.41 & 0.32 & 0.62 & 0.47 & 0.50 & 0.55 & 0.45 & 0.46 & 2.2 \\ 
        \multicolumn{1}{c|}{NIS~\cite{NIS-SLAMZhai}} & 0.30 & 0.40 & 0.36 & 0.29 & 0.31 & 0.92 & 0.67 & 0.44 & 0.46 & \underline{5.0} \\ 
        \multicolumn{1}{c|}{GSS~\cite{matsuki2024gaussian}} & 0.76 & 0.37 & \textbf{0.23} & 0.66 & 0.72 & \underline{0.30} & \textbf{0.19} & 1.46 & 0.58 & 0.7 \\ 
        \multicolumn{1}{c|}{SplaTAM~\cite{keetha2024splatam}} & 0.31 & 0.40 & 0.29 & 0.47 & \underline{0.27} & \textbf{0.29} & 0.32 & 0.55 & \underline{0.36} & 0.2 \\ 
        \multicolumn{1}{c|}{Loopy~\cite{loopyliso}} & \textbf{0.24} & \textbf{0.24} & \underline{0.28} & \underline{0.26} & 0.40 & \textbf{0.29} & \underline{0.22} & \underline{0.35} & \textbf{0.29} & 0.5 \\ 
        \multicolumn{1}{c|}{Ours(Full)} & \underline{0.27}  & \underline{0.28} & 0.32 & \textbf{0.24} & \textbf{0.25} & 0.31 & 0.38 & \textbf{0.30} & \textbf{0.29} & \textbf{11.9}\\ 
        \hline
        \end{tabular}
        \begin{tablenotes}
        \footnotesize
        \item Compared to the tracking results of other NeRF and 3DGS systems. Different From Tab.~\ref{tab:replicaDetail}, we only provide the tracking results here, as some systems are either not open-sourced, not suitable for comparison, or do not provide additional metrics. 
        \end{tablenotes}
        \label{tab:moretracking}
    \end{table}
    
    \noindent
    \textbf{Evaluation on ScanNet \cite{dai2017scannet}.} Although the Replica dataset \cite{straub2019replica} provides accurate reconstructed scenes for evaluating reconstruction metrics, as a synthetic indoor dataset, it does not fully represent the real world. Therefore, we also tested our system on the ScanNet dataset \cite{straub2019replica}, a larger-scale real indoor dataset, which can exhibits the robustness of our system. The tracking results of our system on ScanNet \cite{dai2017scannet} are detailed in Tab. \ref{tab:scannetTrack} and Tab. \ref{tab:scannet}. Each metric consists of the mean and variance obtained by randomly running the system five times on each scene. Our system outperforms other systems in most scenes with a speed of 8 Hz, demonstrating its robustness to real-world environments. Our system also demonstrates higher stability in multiple runs, with much lower standard deviations. 
    
\begin{figure*}[!t]
    \centering
    \begin{minipage}[t]{\textwidth}
        \raggedleft NICE-SLAM \cite{zhu2022nice} \hspace{3.7em} ESLAM \cite{johari2023eslam} \hspace{4.5em} Co-SLAM \cite{wang2023co} \hspace{5.4em} Ours \hspace{7.7em} GT \hspace{4em} \par 
    \end{minipage}
    \vspace{0.5em}
    \begin{minipage}[t]{0.02\textwidth}
    \begin{sideways}
        {\hspace{0.3em} scene0106 \hspace{2em} scene0169 \hspace{3.8em} scene0000 \par  }
    \end{sideways}
    \end{minipage}%
    \begin{minipage}[t]{0.98\textwidth}
      \centering
      \includegraphics[width=\linewidth]{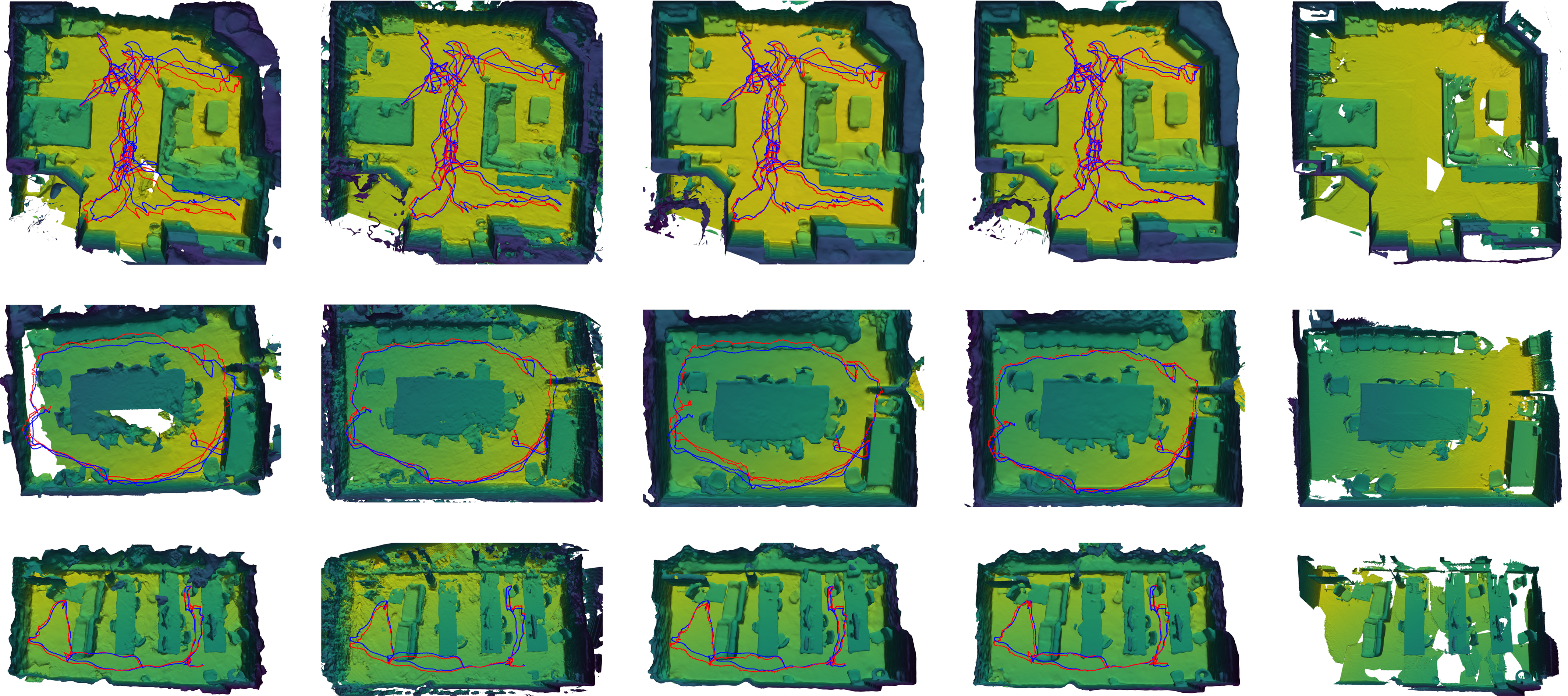} 
    \end{minipage}%
    \hfill
    \caption{Comparison of our system with other state-of-the-art NeRF-based RGB-D dense visual SLAM systems conducted on the ScanNet dataset \cite{dai2017scannet}, focusing on the reconstructed scene depth maps and tracking metrics. Blue lines represent the GT camera trajectory, and red lines indicate the estimated trajectory. Under the same camera perspectives, our system reconstructed overall scene structures more consistent with the GT, along with more accurate camera trajectories.}
    \label{fig:scannet}
    \end{figure*}

\begin{figure*}[h]
        \centering
        \begin{minipage}[t]{\textwidth}
            \raggedleft NICE-SLAM \cite{zhu2022nice} \hspace{3.5em} ESLAM \cite{johari2023eslam} \hspace{4.2em} Co-SLAM \cite{wang2023co} \hspace{5em} Ours \hspace{7.5em} GT \hspace{4em} \par 
        \end{minipage}
        \begin{minipage}[t]{0.02\textwidth}
        \begin{sideways}
            {\hspace{1.6em} scene0169 \hspace{2.5em} scene0106 \hspace{2.5em} scene0000 \par  }
        \end{sideways}
        \end{minipage}%
        \begin{minipage}[t]{0.98\textwidth}
          \centering
          \includegraphics[width=\linewidth]{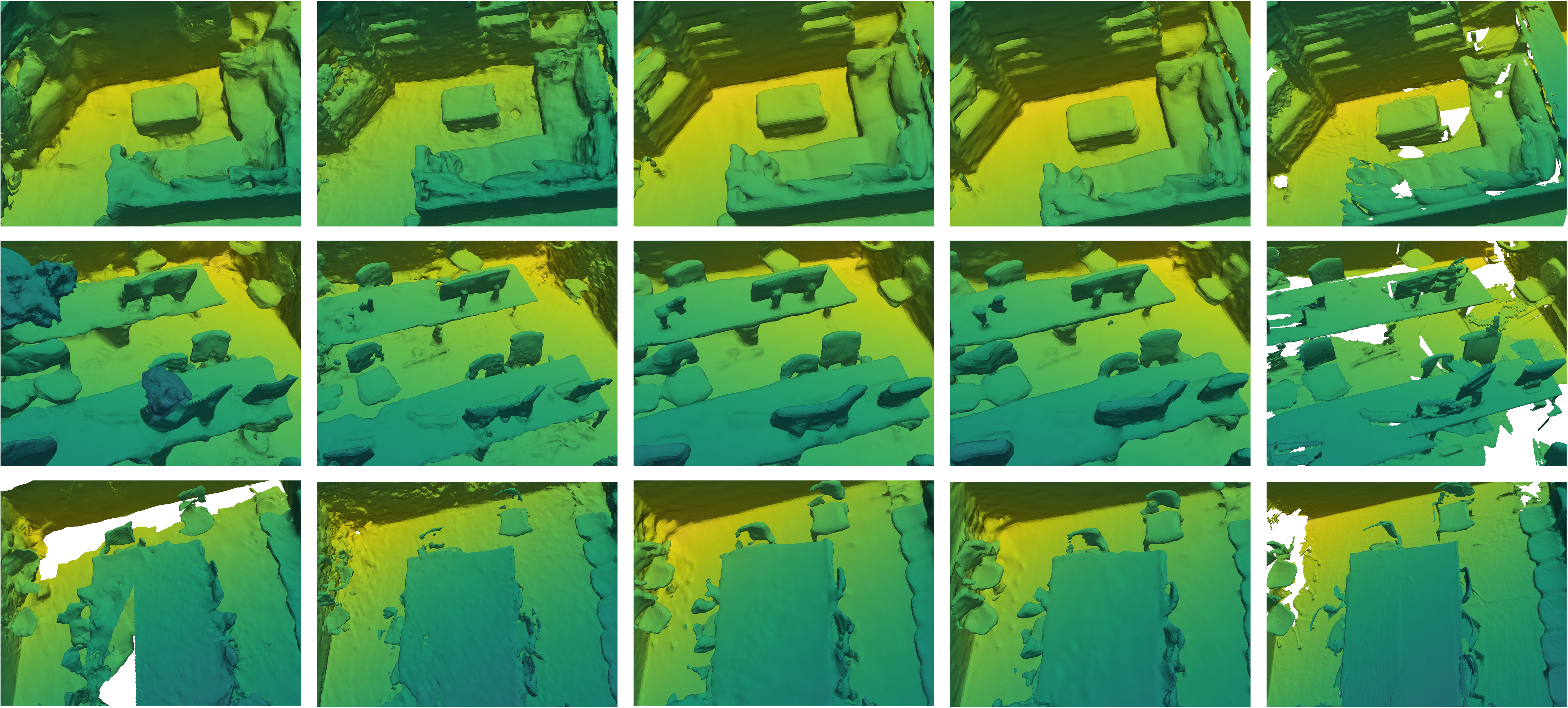} 
        \end{minipage}%
        \hfill
        \caption{Comparison of our system with other state-of-the-art NeRF-based RGB-D dense visual SLAM systems conducted on the ScanNet dataset \cite{dai2017scannet} (scene0000, scene0106, scene0169), focusing on the reconstructed scene depth maps. Under the same camera perspectives, our system reconstructed scene details more consistent with the GT. }
        \label{fig:s2}
    \end{figure*}

    Fig. \ref{fig:scannet} shows our system's reconstruction depth maps and tracking trajectory compared with other baselines on some ScanNet \cite{dai2017scannet} scenes. The tracking results show that our estimated trajectory closely matches the ground truth trajectory. The mapping results indicate that the scenes reconstructed by our system more closely resemble the original scenes, especially in terms of overall structure and shape. For instance, regarding the upper left overall structure of scene 0000, the results computed by our system align more closely with the actual structure. In contrast, the upper left overall structure of the scene generated by other methods is slightly more slanted and larger compared to the Ground Truth. Fig. \ref{fig:s2} presents a comparison of scene detail depth maps under the same camera pose, where our method yields more precise and smoother scene details. For instance, in the scene000 scenario, the positions of various components of the scene displayed by our method under the same camera pose are most similar to those in the ground truth scene.

    \begin{table}[htbp]
        \caption{Tracking Results of Scannet Dataset.}
        \label{tab:scannetTrack}%
        \tabcolsep=0.15cm
        \begin{tabular}{r|ccccccc|c}
        \toprule
        \multicolumn{1}{c|}{Methods}    & 0000 & 0059 & 0106 & 0169 & 0181 & 0207 & Avg. & FPS \\
        \midrule
        \multicolumn{1}{c|}{SNI~\cite{SNISLAMZHU}}   & 6.9 & \underline{7.4} & \underline{7.2} & - & -  & 11.9 & - & -\\
        \multicolumn{1}{c|}{NIS~\cite{NIS-SLAMZhai}}   & 8.7 & 9.6 & 8.4 & \underline{5.6} & - & \underline{7.1} & - & -\\
        \multicolumn{1}{c|}{GSS~\cite{matsuki2024gaussian}}  & 9.6 & \textbf{6.2} & \textbf{7.1} & 10.7 & 18.2 & 7.5 & 9.8 & \underline{2.3}\\ 
        \multicolumn{1}{c|}{SplaTAM~\cite{keetha2024splatam}}  & 12.8 & 10.1 & 17.7 & 12.1 & 11.1 & 7.5 & 11.9 & 0.2 \\
        \multicolumn{1}{c|}{Loopy~\cite{loopyliso}}   & \textbf{4.2} & 7.5 & 8.3 & 7.5 & \underline{10.6} & 7.9 & \underline{7.7} & 0.5\\
        \multicolumn{1}{c|}{Ours}   & \underline{5.3} & 8.2 & 7.3 & \textbf{5.1} & \textbf{8.9} & \textbf{7.0} & \textbf{6.9} & \textbf{8.6} \\
        \bottomrule
        \end{tabular}%
        \begin{tablenotes}
        \footnotesize
        \item Compared to the tracking results of other NeRF and 3DGS systems. Different From Tab.~\ref{tab:scannet}, we only provide the RMSE results here, as some systems are not open-sourced. $-$ indicates that the system is not open-sourced and no results are available.
        \end{tablenotes}
    \end{table}
    \noindent
    \textbf{Evaluation on TUM RGB-D \cite{sturm2012benchmark}. }Consistent with previous literature \cite{zhu2022nice, wang2023co, johari2023eslam}, we further tested our system on three scenes of the TUM dataset. Each result is the average of five random runs of the system. As shown in Tab. \ref{tab:tum}, our system achieved state-of-the-art results in most scenes, further demonstrating our system's cross-dataset robustness. Meanwhile, our speed is the fastest in NeRF based SLAM. Although SLAM systems based on NeRF have lower tracking accuracy compared to traditional SLAM systems, our system significantly narrows this gap with accurate and dense map, providing a robust baseline for future work.

    \label{sec:runtime}

    \begin{table*}[htbp]
    \begin{center}
        \caption{Tracking Results on the Scannet dataset \cite{dai2017scannet}}
        \label{tab:scannet}%
        \tabcolsep=0.3cm
        \begin{tabular}{rc|cccccc|c|c}
        \toprule
        \multicolumn{1}{c}{Methods} & ATE   & Sc. 0000 & Sc. 0059 & Sc. 0106 & Sc. 0169 & Sc. 0181 & Sc. 0207 & Avg. & FPS\\
        \midrule
        \multicolumn{1}{c}{iMAP \cite{sucar2021imap}} & Mean  & 34.2 ± 12.8 & 13.0 ± 2.4 & 12.9 ± 1.7 & 33.6 ± 15.3 & 20.8 ± 3.8 & 18.6 ± 6.0 & 22.2 ± 7.0 & \multicolumn{1}{c}{\multirow{2}[2]{*}{0.1}}\\
              & RMSE  & 42.7 ± 16.6 & 17.8 ± 7.4 & 15.0 ± 1.7 & 39.1 ± 18.2 & 24.7 ± 5.8 & 20.1 ± 6.8 & 26.6 ± 9.4 \\
        \midrule
        \multicolumn{1}{c}{NICE-S \cite{zhu2022nice}} & Mean  & 9.9 ± 0.4 & 11.9 ± 1.8 & 7.0 ± 0.2 & 9.2 ± 1.0 & 12.2 ± 0.3 & 5.5 ± 0.3 & 9.3 ± 0.7 & \multicolumn{1}{c}{\multirow{2}[2]{*}{0.7}}\\
              & RMSE  & 12.0 ± 0.5 & 14.0 ± 1.8 & 7.9 ± 0.2 & 10.9 ± 1.1 & 13.4 ± 0.3 & 6.2 ± 0.4 & 10.7 ± 0.7 \\
        \midrule
        \multicolumn{1}{c}{Co-S \cite{wang2023co}} & Mean  & 6.0 ± 0.1 & 9.4 ± 0.8 & 8.4 ± 0.1 & 5.1 ± 0.1 & 10.3 ± 0.3 & 6.6 ± 0.4 & 7.6 ± 0.3 & \multicolumn{1}{c}{\multirow{2}[2]{*}{6.4}} \\
              & RMSE  & 7.1 ± 0.2 & 11.2 ± 0.7 & 9.3 ± 0.2 & 5.8 ± 0.2 & 11.6 ± 0.6 & 7.1 ± 0.4 & 8.7 ± 0.4 \\
        \midrule
        \multicolumn{1}{c}{E-S \cite{johari2023eslam}} & Mean  & 6.5 ± 0.1 & 6.4 ± 0.4 & 6.7 ± 0.1 & 5.9 ± 0.1 & 8.3 ± 0.2 & \textbf{5.4 ± 0.1} & 6.5 ± 0.2 & \multicolumn{1}{c}{\multirow{2}[2]{*}{4.1}}\\
              & RMSE  & 7.3 ± 0.2 & 8.5 ± 0.5 & 7.5 ± 0.1 & 6.5 ± 0.1 & 9.0 ± 0.2 & \textbf{5.7 ± 0.1} & 7.4 ± 0.2 \\
        \midrule
        \multicolumn{1}{c}{Ours} & Mean  & \textbf{4.7 ± 0.1} & \textbf{5.8 ± 0.2} & \textbf{6.4 ± 0.1} & \textbf{4.6 ± 0.1} & \textbf{8.1 ± 0.1} &  6.6 ± 0.1 & \textbf{6.0 ± 0.1}  & \multicolumn{1}{c}{\multirow{2}[2]{*}{\textbf{8.6}}}\\
              & RMSE  & \textbf{5.3 ± 0.2} & \textbf{8.2 ± 0.4} & \textbf{7.3 ± 0.1} & \textbf{5.1 ± 0.1} & \textbf{8.9 ± 0.2} &  7.0 ± 0.1 & \textbf{6.9 ± 0.2} \\
        \bottomrule
        \end{tabular}%
        \begin{tablenotes}
        \footnotesize
        Quantitative comparison of our system against other SOTA NeRF-based RGB-D dense visual SLAM systems, using the ScanNet dataset \cite{dai2017scannet}. We get the results of ESLAM \cite{johari2023eslam}, Nice-SLAM \cite{zhu2022nice} and iMap \cite{sucar2021imap} from ESLAM \cite{johari2023eslam}. Numbers in \textbf{bold black font} represent the best results in each column. Our tracking results outperformed those of other systems in most scenes, with the fastest FPS.
        \end{tablenotes}
    \end{center}
    \end{table*}%
    
\subsection{Implicit loop closure capabilities of NeRF}

\begin{figure}[h]
        \centering
        \subfloat[{\small The frame-wise quaternion discrepancies.}]{\includegraphics[width=\linewidth]{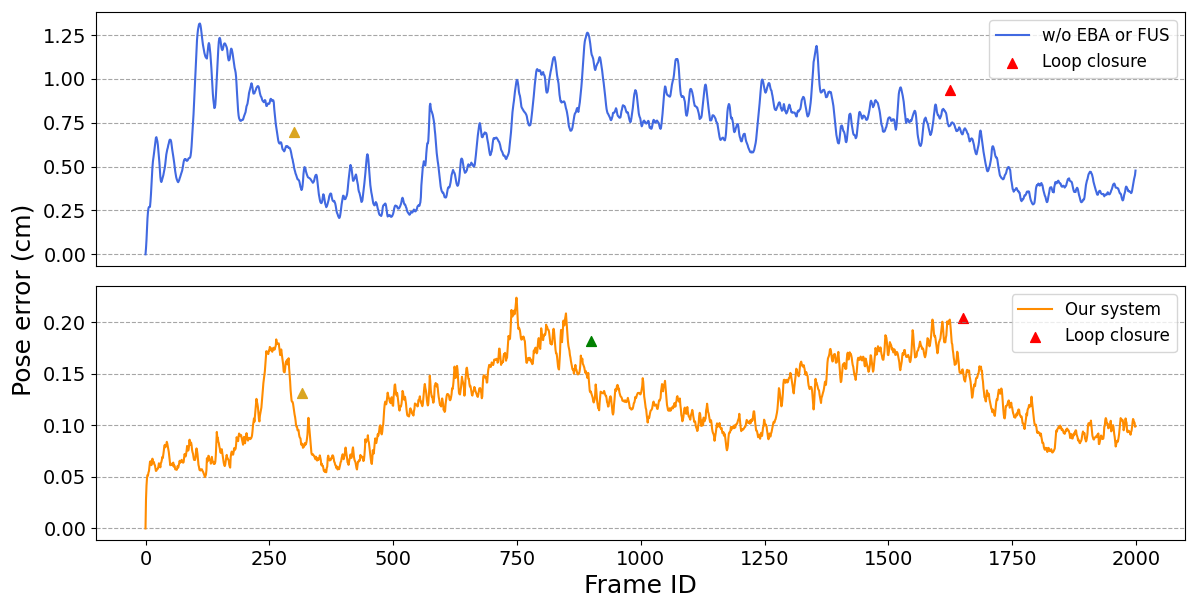}}
        
        \subfloat[\small Detailed loop closure location]{\includegraphics[width=\linewidth]{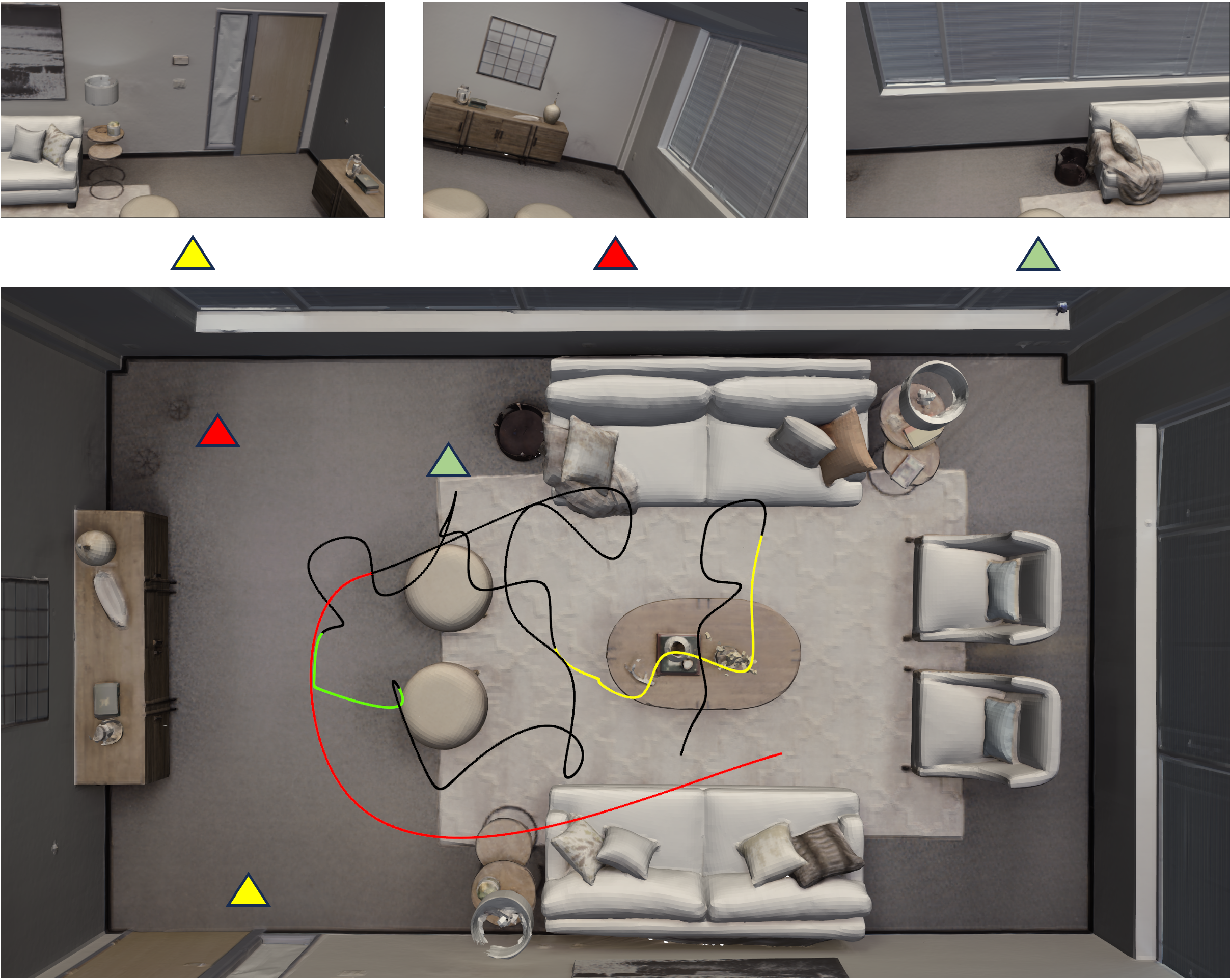}}
        \caption{The implicit loop closure capabilities of NeRF on the Replica dataset: (a) The upper figure shows the results of frame-by-frame pose quaternion differences without FUS and EBA, while the lower figure presents the results with FUS and EBA. Different colored solid triangles represent the locations of implicit loop closures. (b) Shows the actual areas of loop closures corresponding to figure (a), with colored solid triangles matching the three loop closure colors in figure (a). The black trajectory corresponds to the actual trajectory, with other colored trajectories indicating the locations of the three loop closures identified in figure (a).} 
        \label{fig:loop}
    \end{figure}

    Our observations reveal that NeRF intrinsically supports implicit loop closure correction. This capability is illustrated in the upper figure of Fig.~\ref{fig:loop}(a), using the Replica dataset's \textit{Room0} scene, in the absence of feature-based and uniform sampling (FUS) or effective constraint through global bundle adjustment (EBA). Notably, at frames 270 and 1625, a significant reduction in error is observed, corresponding with the system revisiting previously encountered locations (Fig.~\ref{fig:loop}(b)). Our observations of implicit loop closure capabilities in NeRF are not isolated to a single scene or dataset. We consistently observe similar phenomena across multiple scenes in various datasets, which further corroborates NeRF's proficiency in implicit loop closure. Upon integrating FUS along with EBA, there is a notable reduction in the overall pose error of the system. Additionally, this integration facilitated the identification of an extra loop closure event at frame 875, as depicted in the lower figure of Figure~\ref{fig:loop}(a). This event serves as compelling evidence of NeRF’s implicit loop closure capabilities.
    
    The observed capability for implicit loop closure in NeRF stems from its continuous and dense spatial representation, while our FUS sampling further enhances this capability. The optimized implicit mapping ability, to some extent, resembles the biological perception pattern of the environment: the brain creates a dense map of the space while focusing on recording some meaningful objects (such as tables and chairs). When the eyes perceive a scene, the brain continuously compares it with existing scenes in the mind, relying on certain feature objects (attention-based) for comprehensive localization. Compared to the explicit recognition of loop closures in classic SLAM, NeRF's implicit loop closure appears more natural because it can comprehensively consider both the continuity and specificity of space. It demonstrates the powerful spatial representation capabilities of NeRF.

    \begin{table}[htbp]
    \begin{center}
        \caption{Tracking Results on the Tum dataset \cite{sturm2012benchmark}}
        \label{tab:tum}%
        \tabcolsep=0.2cm
        \begin{tabular}{c|c|ccc|c}
        \toprule
           &Method   & fr1/desk & fr2/xyz & fr3/office & FPS \\
        \midrule
        \multirow{3}[2]{*}{\begin{sideways} Classic \end{sideways}} 
        &Kintinuous \cite{newcombe2011kinectfusion} & 3.70  & 2.90  & 3.00 & \textbf{40} \\
        &BAD-SLAM \cite{schops2019bad} & 1.70  & 1.10  & 1.70 & 2 \\
        &ORB-SLAM2 \cite{mur2017orb} & \textbf{1.60}  & \textbf{0.40} & \textbf{1.00} & 30\\
        \midrule
        \multirow{6}[2]{*}{\begin{sideways} NeRF \end{sideways}} 
        &iMAP \cite{sucar2021imap} & 4.90  & 2.00  & 5.80 & 0.1 \\
        &NICE-SLAM \cite{zhu2022nice} & 2.85  & 1.84  & 2.95 & 0.1 \\
        &Co-SLAM \cite{wang2023co} & 2.44  & 1.71 & 2.46 & 6.9 \\
        &E-SLAM \cite{johari2023eslam} & 2.54  & \textbf{1.09}  & 2.47 & 0.2 \\
        &Loopy~\cite{loopyliso} & 3.79 & 1.62 & 3.41 & 0.5\\
        &Ours & \textbf{2.32}  & 1.90  & \textbf{2.41} & \textbf{7.1} \\
        \bottomrule
        \end{tabular}%
        \begin{tablenotes}
        \footnotesize
        Numbers in \textbf{bold black font} represent the best results in each column. Our method surpasses the accuracy of other NeRF-based methods and narrows the gap with traditional approaches.
        \end{tablenotes}
    \end{center}
    \end{table}%
    
\subsection{Runtime Analysis}

       \begin{table}[htbp]
    \begin{center}
        \caption{Time and Memory Analysis.}
        \label{tab:timeEfficiency}%
        \tabcolsep=0.4cm
        \begin{tabular}{c|c|c|c}
        \toprule
            & \multicolumn{1}{c|}{Method} & FPS   & \multicolumn{1}{c}{Model Param.} \\
        \midrule
        \multirow{7}[2]{*}{\begin{sideways}Replica\end{sideways}} 
              & NICE-SLAM \cite{zhu2022nice} & 0.9  & 49.0 M  \\
              & Co-SLAM \cite{wang2023co} & 17.1  & \textbf{7.9} M      \\
              & ESLAM \cite{johari2023eslam} & 12.1  & 23.2 M  \\
              & GSS \cite{matsuki2024gaussian} & 0.7 & 24.2 M \\
              & SplaTAM \cite{keetha2024splatam} & 0.2 & 243.4 M \\
              & Ours (Lite)  & \textbf{21.4}  & 8.2 M   \\
              & Ours (Full)  & 11.9  & 8.2 M   \\
        \midrule
        \multirow{6}[2]{*}{\begin{sideways}ScanNet\end{sideways}} 
              & NICE-SLAM \cite{zhu2022nice} & 0.7  & 88.7 M  \\
              & Co-SLAM \cite{wang2023co} & 6.4 & 9.9 M       \\
              & ESLAM \cite{johari2023eslam} & 4.1  & 70.9 M  \\
              & GSS \cite{matsuki2024gaussian} & 2.3 & \textbf{5.6} M \\
              & SplaTAM \cite{keetha2024splatam} & 0.2 & 156.4 M \\
              & Ours & \textbf{8.6}  & 11.3 M \\
        \bottomrule
        \end{tabular}%
        \begin{tablenotes}
        \footnotesize
        Comparison of the model size and runtime for maps generated by different systems on the Replica \cite{straub2019replica} and ScanNet \cite{dai2017scannet} datasets. Numbers in \textbf{bold black font} represent the best results in each column.
        \end{tablenotes}
    \end{center}
    \end{table}%
    
    We evaluated the model size and FPS produced by all systems on the Replica \cite{straub2019replica}, and ScanNet \cite{dai2017scannet} datasets. For the Replica dataset \cite{straub2019replica}, we tested the parameter count and FPS for room0, and for the ScanNet dataset \cite{dai2017scannet}, we assessed the parameter count and runtime for scene0000. The details are provided in Tab. \ref{tab:timeEfficiency}. All results were obtained on the same computer configuration. The results indicate that our system operates faster than others, with a small model size as Co-SLAM \cite{wang2023co}, attributed to using a hash sparse grid representation of the map and our one-time sampling iteration optimization strategy.

\subsection{Ablations}

    We conducted ablation experiments on our effectively constrained global BA strategy and feature-based and uniform sampling, detailed in Tab. \ref{tab:ablation}. There are four comparative experiments: one without the effectively constrained global BA strategy, one without feature-based and uniform sampling, one with neither, and one with both. Without the effectively constrained global BA strategy, we randomly selected the number and order of keyframes in the sliding window and the number of sampled pixels per keyframe. Without feature-based and uniform sampling, we randomly sampled pixels. The rest of the parameters remained the same in all experiments except for the listed variable differences. The results show that both optimization strategies significantly reduce our tracking error and improve our mapping accuracy. It is noted that FUS contributes to a more significant improvement in system accuracy compared to EBA, as it markedly reduces randomness and fully leverages the implicit loop closure capabilities of NeRF.

    \begin{table}[htbp]
    \begin{center}
      \caption{Ablation study.}
      \label{tab:ablation}%
      \centering
      \tabcolsep=0.35cm
        \begin{tabular}{ccccc}
        \toprule
        Name  & FUS   & EBA  & RMSE.(cm) & Mean.(cm) \\
        \midrule
        w/o both     &       &       &  6.9 ± 0.2 & 6.1 ± 0.1 \\
        w/o EBA     & \checkmark     &       &  6.4 ± 0.2 & 5.5 ± 0.2 \\ 
        w/o FUS     &       & \checkmark     & 5.7 ± 0.2 & 5.1 ± 0.2 \\
        Full model & \checkmark     & \checkmark     & \textbf{5.3 ± 0.2}  & \textbf{4.7 ± 0.1} \\
        \bottomrule
        \end{tabular}%
        \begin{tablenotes}
        \footnotesize
        Ablation of BA strategies on our system on scene0000 of the ScanNet dataset \cite{dai2017scannet}: EBA represents our effectively constrained global Bundle Adjustment strategy; FUS represents our feature-based and uniform sampling. Numbers in \textbf{bold black font} represent the best results in each column. 
       \end{tablenotes}
    \end{center}
    \end{table}%

\section{Conclusion}
\label{sec:cons}


We present EC-SLAM, a real-time, NeRF-based RGB-D SLAM system that delivers exceptional accuracy and efficiency. Our innovations include an effectively constrained global bundle adjustment strategy, feature-based and uniform sampling, sparse positional encoding, and a TSDF-based loss function. These enhancements fully exploit NeRF's implicit loop closure capabilities, resulting in precise camera trajectory tracking and map reconstruction with minimal model parameters. Extensive experiments on multiple datasets demonstrate EC-SLAM's state-of-the-art performance and impressive 21 Hz runtime.

While EC-SLAM offers significant advancements, opportunities for improvement remain. Our future work concentrates on extending the capabilities of EC-SLAM. We will explore multi-sensor integration to boost robustness in diverse environments, investigate combining classic SLAM algorithms with NeRF-based methods for enhanced performance, and work towards adapting EC-SLAM to tackle the complexities of large-scale outdoor exploration. These advancements will drive the development of even more powerful and versatile SLAM systems.



{
    \bibliographystyle{IEEEtran}
    \bibliography{main}
}

\begin{IEEEbiography}[{\includegraphics[width=1in,height=1.25in,clip,keepaspectratio]{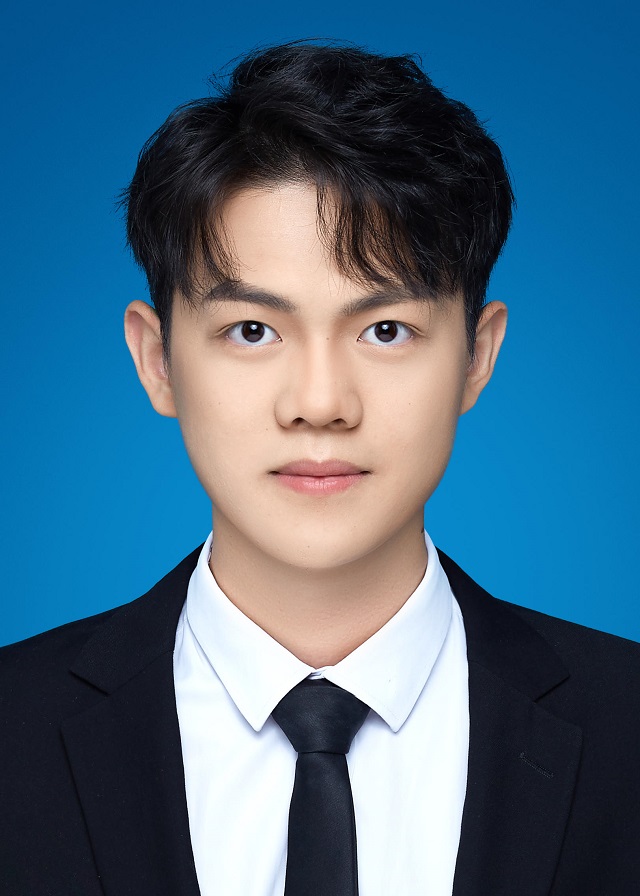}}]{Guanghao Li} received the B.S. degree from Xiamen University. He is currently pursuing his Ph.D. at the Institute of Science and Technology for Brain-Inspired Intelligence (ISTBI), Fudan University. His research areas are focused on Visual SLAM and Neural Radiance Fields (NeRF). 
\end{IEEEbiography}

\begin{IEEEbiography}[{\includegraphics[width=1in,height=1.25in, clip,keepaspectratio]{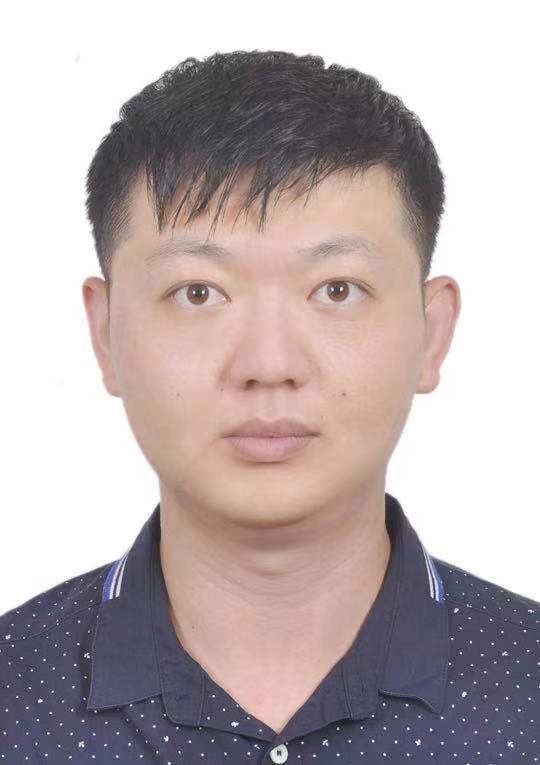}}]{Qi Chen} received a B.S. degree from the Beijing Institute of Technology in 2016 and an M.S. degree from New York University in 2018. From 2019 to 2021, he was an algorithm engineer at the Institute of Acoustics of the Chinese Academy of Sciences. He is currently a Ph.D. student in computer science at Fudan University, Shanghai, China. His research interests include computer vision, digital signal processing, and machine learning. 
\end{IEEEbiography}

\begin{IEEEbiography}[{\includegraphics[width=1in,height=1.25in, clip,keepaspectratio]{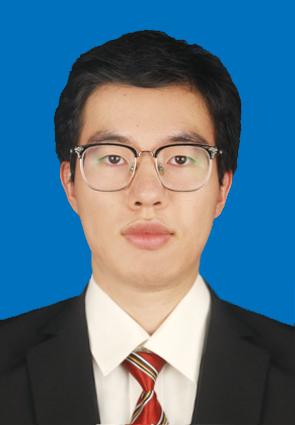}}]{Yuxiang Yan} received a B.S. degree from East China University of Science and Technology. He is currently pursuing his Ph.D. at the Institute of Science and Technology for Brain-Inspired Intelligence, Fudan University, Shanghai, China. His current research areas are focused on computer vision and machine learning for autonomous driving.
\end{IEEEbiography}

\begin{IEEEbiography}[{\includegraphics[width=1in,height=1.25in,clip,keepaspectratio]{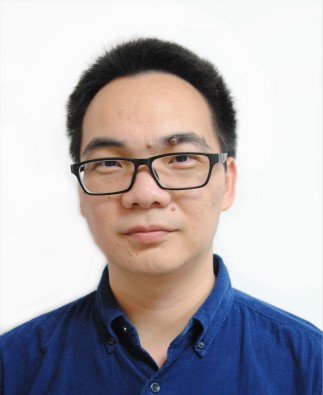}}]{Jian Pu} received a Ph.D. degree from Fudan University, Shanghai, China, in 2014. Currently, he is a Young Principal Investigator at the Institute of Science and Technology for Brain-Inspired Intelligence (ISTBI), Fudan University. He was an associate professor at the School of Computer Science and Software Engineering, East China Normal University from 2016 to 2019, and a postdoctoral researcher of the Institute of Neuroscience, Chinese Academy of Sciences in China from 2014 to 2016. His current research interest is to develop machine learning and computer vision methods for autonomous driving.
\end{IEEEbiography}

\end{document}